\documentclass[conference]{IEEEtran}
\IEEEoverridecommandlockouts

\usepackage{cite}
\usepackage{amsmath,amssymb,amsfonts}
\usepackage{algorithmic}
\usepackage{commath}
\usepackage{nicematrix}
\usepackage{graphicx}
\usepackage{enumitem}
\usepackage{booktabs}
\usepackage{textcomp}
\usepackage{xcolor}
\def\BibTeX{{\rm B\kern-.05em{\sc i\kern-.025em b}\kern-.08em
    T\kern-.1667em\lower.7ex\hbox{E}\kern-.125emX}}
\usepackage[pagebackref,breaklinks,colorlinks]{hyperref}
\usepackage{fancyhdr}
\fancypagestyle{firststyle}
{
   \fancyhf{}
   \fancyhead[C]{\footnotesize Accepted at the European Solid-state Devices and Circuits Conference (ESSDERC), September 2022}
}

\begin{document}
\bstctlcite{IEEEexample:BSTcontrol}

\title{In-memory Realization of In-situ \\ Few-shot Continual Learning with \\ a Dynamically Evolving Explicit Memory
}

\author{\IEEEauthorblockN{G. Karunaratne\IEEEauthorrefmark{1}\IEEEauthorrefmark{4},
M. Hersche\IEEEauthorrefmark{1}\IEEEauthorrefmark{4}, 
J. Langenegger\IEEEauthorrefmark{1}\IEEEauthorrefmark{4},
G. Cherubini\IEEEauthorrefmark{1}, 
M. Le Gallo\IEEEauthorrefmark{1},
U. Egger\IEEEauthorrefmark{1},\\
K. Brew\IEEEauthorrefmark{2},
S. Choi\IEEEauthorrefmark{2},
I. Ok\IEEEauthorrefmark{2},
C. Silvestre\IEEEauthorrefmark{2},
N. Li\IEEEauthorrefmark{2},
N. Saulnier\IEEEauthorrefmark{2},
V. Chan\IEEEauthorrefmark{2},
I. Ahsan\IEEEauthorrefmark{2},\\
V. Narayanan\IEEEauthorrefmark{3},
L. Benini\IEEEauthorrefmark{4},
A. Sebastian\IEEEauthorrefmark{1},
A. Rahimi\IEEEauthorrefmark{1}}\\
\vspace{-0.3cm}
\IEEEauthorblockA{\IEEEauthorrefmark{1}IBM Research,
Z\"{u}rich, Switzerland 
\IEEEauthorrefmark{2}IBM Research, Albany, NY, USA\\
\IEEEauthorrefmark{3}IBM T. J. Watson Research Center, NY, USA
\IEEEauthorrefmark{4}ETH Z\"{u}rich, Z\"{u}rich, Switzerland}\\}

\maketitle
\thispagestyle{firststyle}
\begin{abstract}
Continually learning new classes from few training examples without forgetting previous old classes demands a flexible architecture with an inevitably growing portion of storage, in which new examples and classes can be incrementally stored and efficiently retrieved. One viable architectural solution is to tightly couple a stationary deep neural network to a dynamically evolving explicit memory (EM). As the centerpiece of this architecture, we propose an EM unit that leverages energy-efficient in-memory compute (IMC) cores during the course of continual learning operations. We demonstrate for the first time how the EM unit can physically superpose multiple training examples, expand to accommodate unseen classes, and perform similarity search during inference, using operations on an IMC core based on phase-change memory (PCM). Specifically, the physical superposition of few encoded training examples is realized via in-situ progressive crystallization of PCM devices. The classification accuracy achieved on the IMC core remains within a range of 1.28\%--2.5\% compared to that of the state-of-the-art full-precision baseline software model on both the CIFAR-100 and miniImageNet datasets when continually learning 40 novel classes (from only five examples per class) on top of 60 old classes. 
\end{abstract}

\begin{IEEEkeywords}
In-memory Computing, Continual Learning, Few-shot Learning, Hyperdimensional Computing, Non-volatile Memory Devices
\end{IEEEkeywords}

\section{Introduction}
Few-shot continual learning (FSCL), aka few-shot class-incremental learning~\cite{FSCIL_CVPR2020,shi_nips2021}, requires a learner to incrementally learn new classes from very few training examples, without forgetting the previously learned classes. The learner is exposed to a series of sessions, whereby each session introduces distinct unseen classes by providing only a few training examples per class. From these few examples, the learner should quickly and incrementally learn novel classes without forgetting the previously learned old classes. After learning novel classes in each session, the learner is evaluated on several query samples from all the classes, to which it was exposed so far (i.e., the union of the classes from the previous and the current sessions). FSCL is a very challenging research problem that could impose significant additional compute and memory costs on the learner.

Very recently, a low-cost solution was proposed that avoids expensive gradient-based computations for learning unseen classes during the course of FSCL~\cite{C-FSCIL_CVPR22}. This solution is inspired by a robust few-shot learner~\cite{kar_ncom_2021} that brings together deep neural networks with hyperdimensional computing~\cite{Kanerva2009,VSA03} to be able to represent raw images with high-dimensional holographic binary or bipolar vectors. Inspired by this powerful combination, the learner in~\cite{C-FSCIL_CVPR22} is composed of a \emph{frozen} controller and a \emph{dynamically evolving} explicit memory (EM) for FSCL. The controller is a deep convolutional network (including a final fully connected layer) that is interfaced with the EM unit to dynamically store or retrieve the acquired knowledge about the classes. The controller interacts with the EM through write and read operations using $d$-dimensional holographic vectors. Although the controller remains stationary, the EM dynamically updates its contents by new examples, and grows its size by storing new classes. Hence, it is desirable to have an EM unit with dynamically evolving contents that retains the acquired knowledge about already-seen classes in both compressed and nonvolatile manner. 

\begin{figure}[!ht]
\centering
\includegraphics[width=0.48\textwidth]{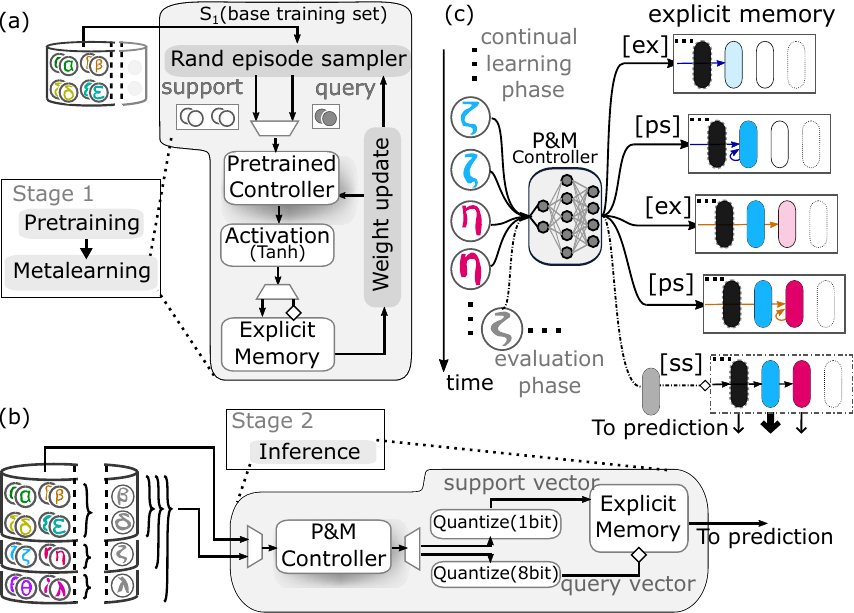}
\caption{\textbf{(a)} Stage 1 (Pretraining and Metalearning), the controller weights are updated over a series of episodes. \textbf{(b)} In Stage 2 (Inference), the pretrained and metalearned (P\&M) controller generates support and query vectors over a series of sessions. \textbf{(c)} EM operations during inference stage can be divided into two phases. Phase 1 continual learning consists of: [ex]=expansion (allocating a new column in the array to store the first support vector from a previously unseen class); [ps]=physical superposition (few support examples are accumulated on the relevant class memory). Phase 2 evaluation phase consists of: [ss]=similarity search between a query vector and class vectors on the array at that point in time. Black and white slots in EM correspond to already filled and empty slots, respectively.} \label{fig:phases}
\end{figure}

\section{Proposed Explicit Memory Unit for FSCL}
Fig.~\ref{fig:phases}(a)(b) illustrates the main stages of FSCL involved in~\cite{C-FSCIL_CVPR22}: a pretraining and metalearning stage, and an inference stage. The first stage is all done in software. The goal of this rather elaborate stage is to train the controller (here, a ResNet-12) to be able to generate $d$-dimensional\footnote{As seen later in Section~\ref{sec:setup}, $d$ is fixed to 256 so that the vectors occupy an entire column of a PCM crossbar array.} quasi-orthogonal real-valued vectors for different classes. By the end of this stage, the controller has learned how to assign 256-dimensional quasi-orthogonal, and thus dissimilar, vectors to novel classes in the EM, which allows the controller to remain stationary afterwards. This pretraining and metalearning stage is based on just the first session's (S1) training samples. Based on the datasets used in our experiments, as explained in Section~\ref{sec:dataset}, the first session contains 60 classes, each including 500 samples, out of the total 100 classes. 

Next, the inference stage is composed of two phases, as shown in Fig.~\ref{fig:phases}(c): a continual learning phase of novel classes from very few training/support examples per class (in our results, 5 training examples, hence 5-shot), and a query evaluation phase in which we evaluate the accuracy over a batch of query examples containing a number of samples (100 in the datasets we used) per each class encountered so far. We quantize the controller’s output vector elements to 1-bit, hence the controller generates bipolar support vectors (of training examples) to be written, or accumulated, in the EM unit during the continual learning phase. For the query vectors (of query samples) during the evaluation phase, we quantize the controller’s output vector elements to 8-bit, so an analog input corresponding to the 8-bit query vector element is applied in each row of the crossbar array. This performs a similarity search between query vector and the class vectors. The resulting vector received via the columns of the crossbar array is used to classify the query sample. The class corresponding to the maximum element of the resulting vector is then taken as the predicted class. 

\begin{figure}[!ht]
\includegraphics[width=0.48\textwidth]{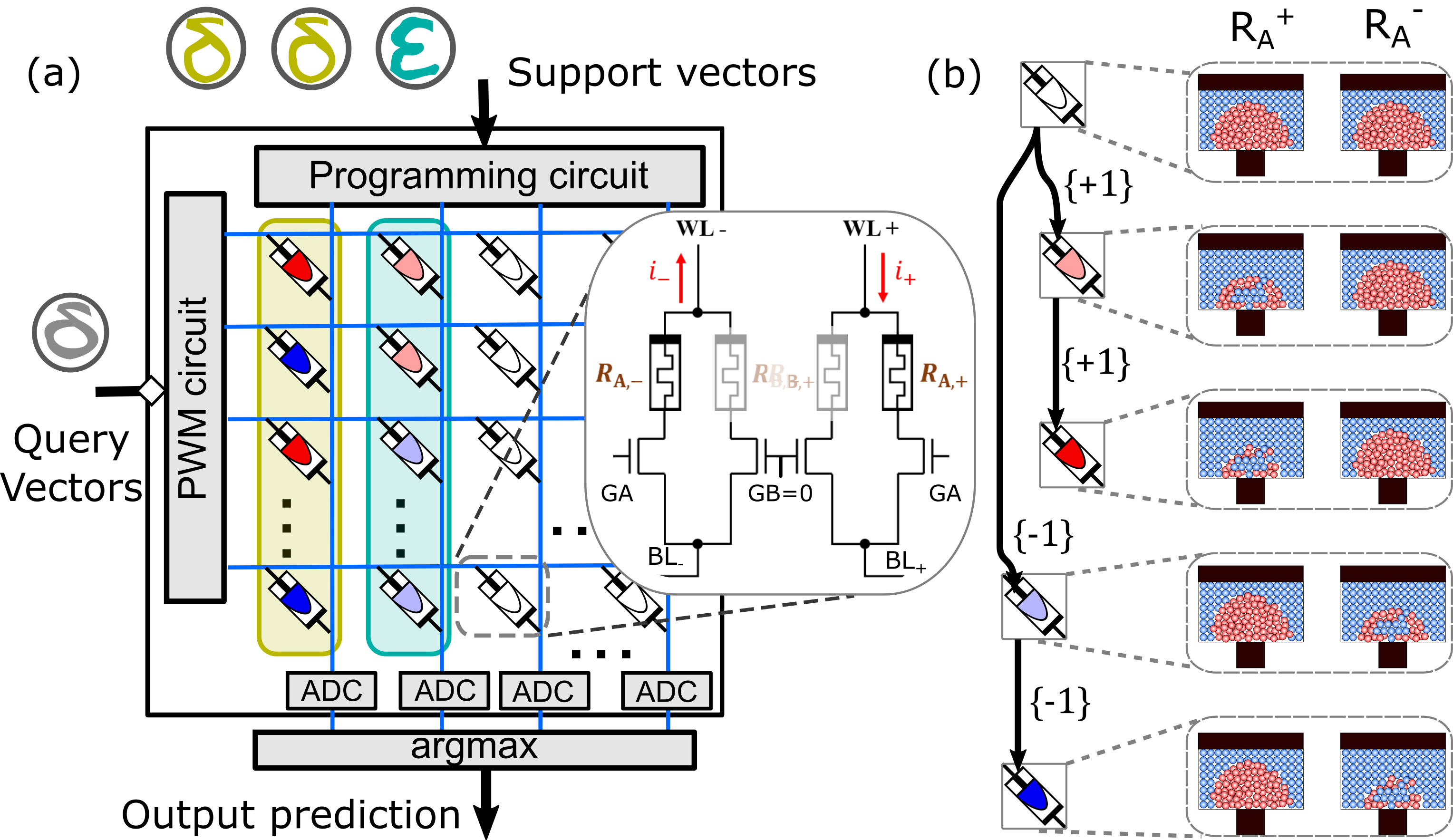}
\caption{\textbf{(a)} Schematic illustration of the EM implemented on a cross-bar array of unit-cells. Each unit-cell comprises 4 PCM devices (only 2 used for this experiment) organized in a differential configuration. During continual learning, the expansion and physical superposition operations are performed by applying SET pulses to a column of devices. Similarity search is realized by applying read voltage pulses with varying pulse width and subsequently digitizing the accumulated current using ADCs. \textbf{(b)} Illustration of the partial crystallization of the differential pair of devices under different +1/-1 pulse sequences during superposition operation.} \label{fig:arch}
\end{figure}

For the inference stage, the EM is implemented on an IMC core with a unit-cell array comprising PCM devices (see Fig.~\ref{fig:arch}). In every continual learning phase: (i) when the first example of a new class appears, the EM is expanded by choosing a fully reset column of unit-cells and, based on the sign of the bipolar support vector element, the unit-cell conductance is increased or decreased by the application of single SET pulses; ii) when an example from a previously seen class appears, a similar update is performed on the column of unit cells corresponding to that particular class.

We exploit the in-situ accumulation via progressive crystallization of PCM to realize physical superposition of few related support vectors on a single class vector. This means, as shown in Fig.~\ref{fig:arch}(b), starting from a fully reset pair of PCM devices, fine grained SET pulses are applied on either the positive or the negative device depending on whether the type of accumulation is incremental or decremental based on the input bipolar support vector element. This creates a multibit analog EM, whereby the size of the EM is set just by the number of classes, as opposed to the product of the number of classes and the number of training examples per class~\cite{kar_ncom_2021}. The nonvolatility of the resulting analog states preserves the acquired knowledge about already-seen classes. Finally, during the evaluation phase, the frequent similarity search between the 8-bit query vector (of a query image) and the analog class vectors are computed in-memory by exploiting Kirchhoff’s circuit laws, and the result is directly used for classification. 

Moreover, given that the controller requires no parameter updates after the first stage of pretraining and metalearning, it can be treated as a deep neural network with stationary weights that can also benefit from an IMC-based implementation, as demonstrated for example in~\cite{joshi_ncom2020}.

\begin{figure}[!ht]
\includegraphics[width=0.48\textwidth]{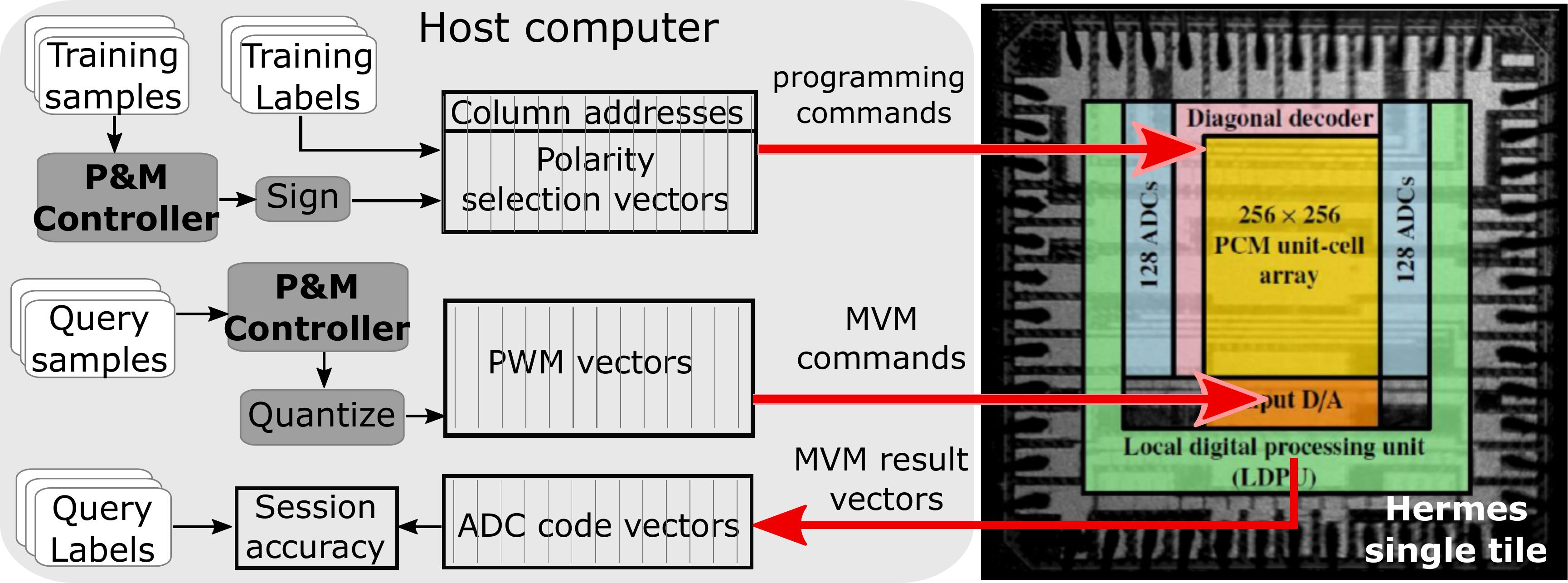}
\caption{Experimental Setup. On the right: the micrograph of the Hermes chip. The chip is accessed using a sequence of programming and MVM commands prepared and sent by the host computer via a FPGA-based interface.} \label{fig:exp_setup}
\end{figure}

\begin{figure}[!ht]
\includegraphics[width=0.48\textwidth]{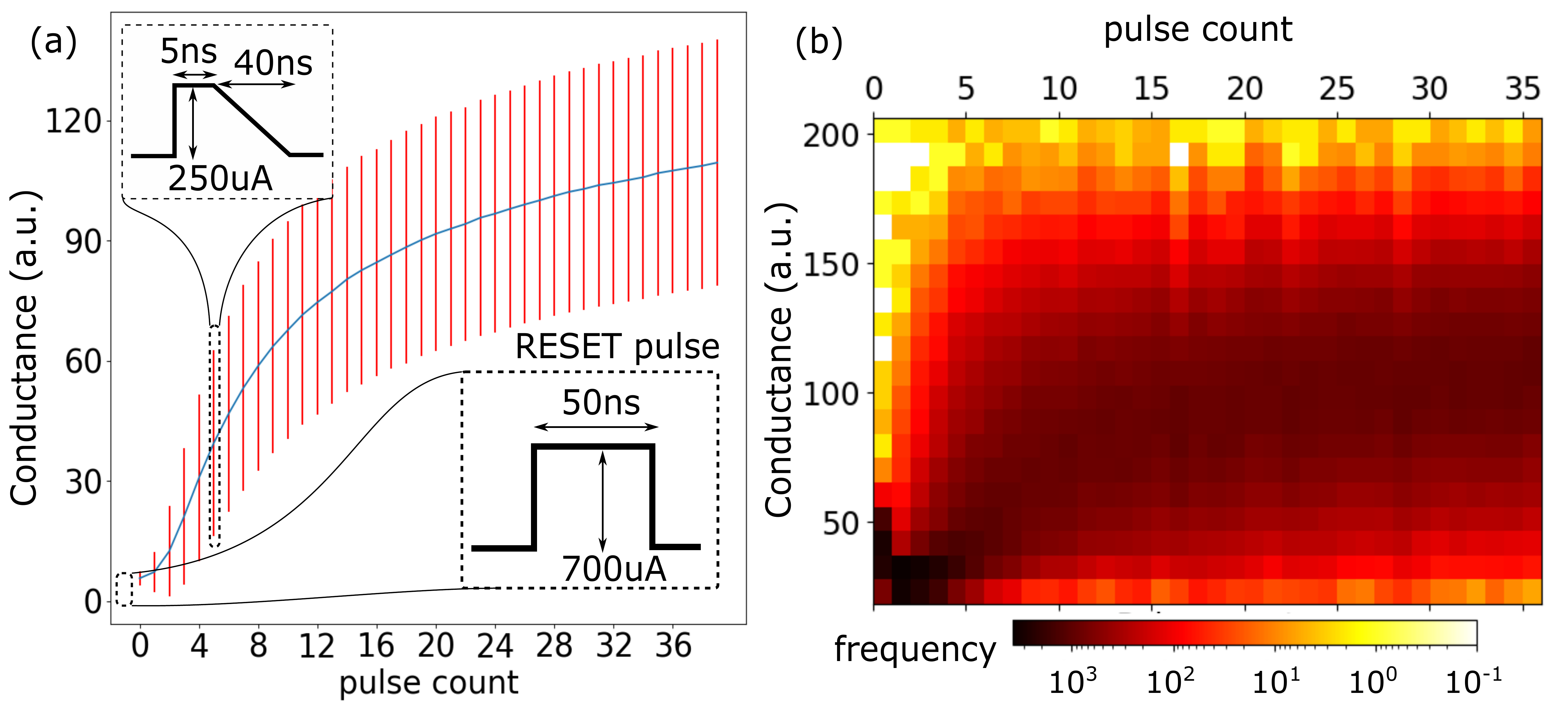}
\caption{\textbf{(a)} Progressive crystallization of PCM devices in the IMC core. The conductance distribution corresponding to 65,536 devices is shown with the successive application of SET pulses after an initial RESET. The blue curve denotes the mean conductance and the red bars indicate one standard deviation. \textbf{(b)} A 2-D illustration of the conductance distribution for the same experiment. It can be seen that the dynamic range (number of SET pules needed to span the conductance range) is fully adequate, because of the very small number (~5) of training examples available during FSCL.}
\label{fig:programming}
\end{figure} 


\section{Experimental Setup}
\label{sec:setup}
The experiments are carried out on the Hermes chip with a 256x256 unit-cell array of PCM devices organized in a differential configuration \cite{Y2022khaddamJSSC} (see Fig.~\ref{fig:exp_setup}), accessed by a host computer via FPGA (Field Programmable Gate Array)-based interface. The bipolar support vectors are sent as SET pulses to the corresponding column of 256 unit cells of the initially fully RESET array. Based on the +1/-1 vector elements, the conductance of the positive/negative polarity devices are increased. The evolution of the conductance distribution of individual PCM devices as a function of the number of applied SET pulses on initially RESET devices is shown in Fig.~\ref{fig:programming}.

For similarity search in the evaluation phase, a 4-quadrant matrix-vector multiply (MVM) is performed between the 8-bit query vector and the set of analog class vectors stored in the unit-cell array. 

\begin{figure}[!ht]
\includegraphics[width=0.48\textwidth]{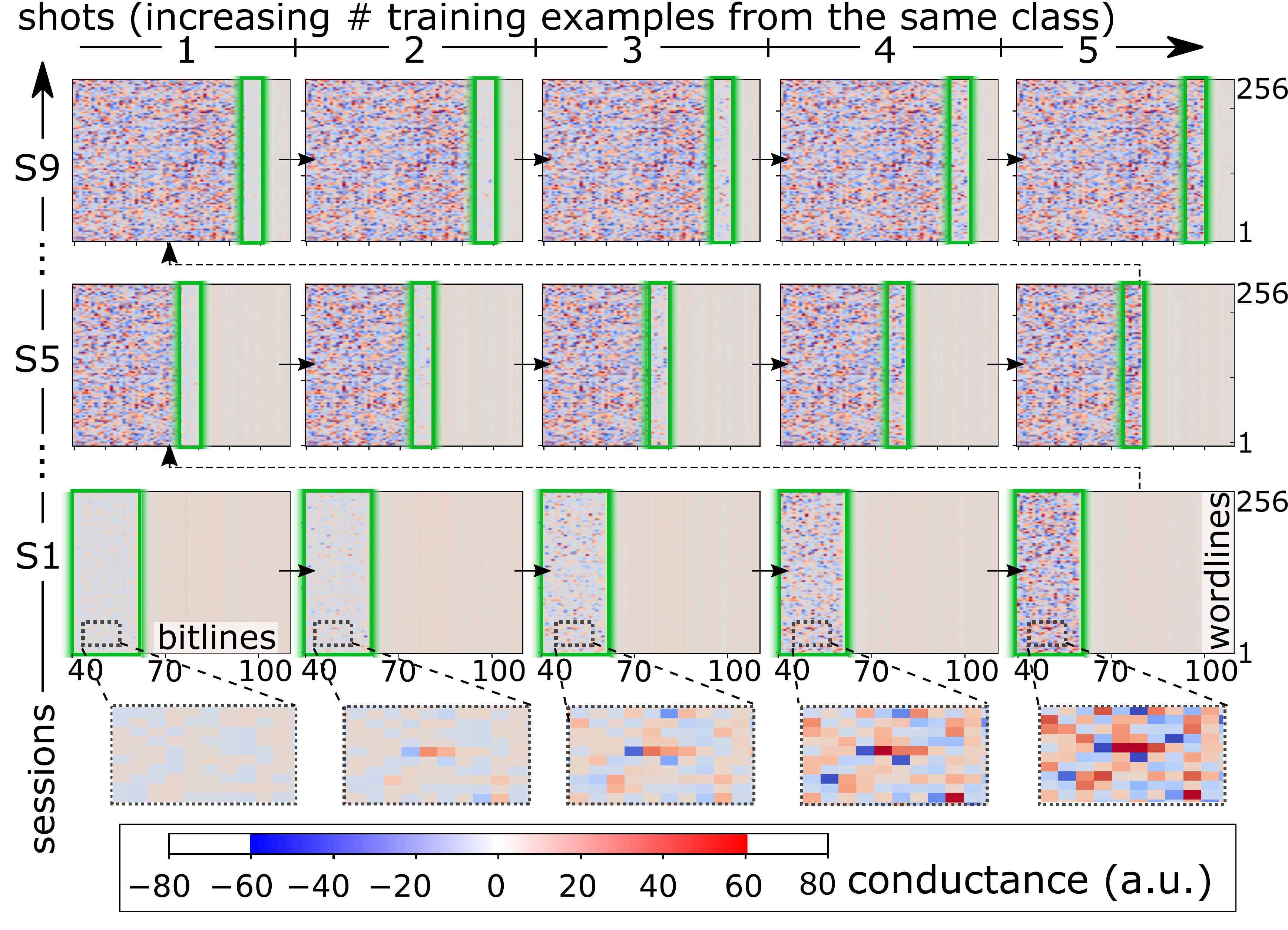}
\caption{A 2-D map of the array conductance (of the unit-cells with differential PCM devices) for the FSCL experiment on CIFAR-100 dataset. With each new session, more columns of the array are selected corresponding to the new classes in that session. And with each training example per class, the corresponding class vectors are updated with the application of SET pulses. The unit-cells activated for the update at each panel are highlighted in green. A close up view of the conductance evolution in a 10x10 region of the crossbar is shown in the bottom row.}
\label{fig:condmaps2}
\end{figure}

\begin{figure}[!ht]
\centering
\includegraphics[width=0.48\textwidth]{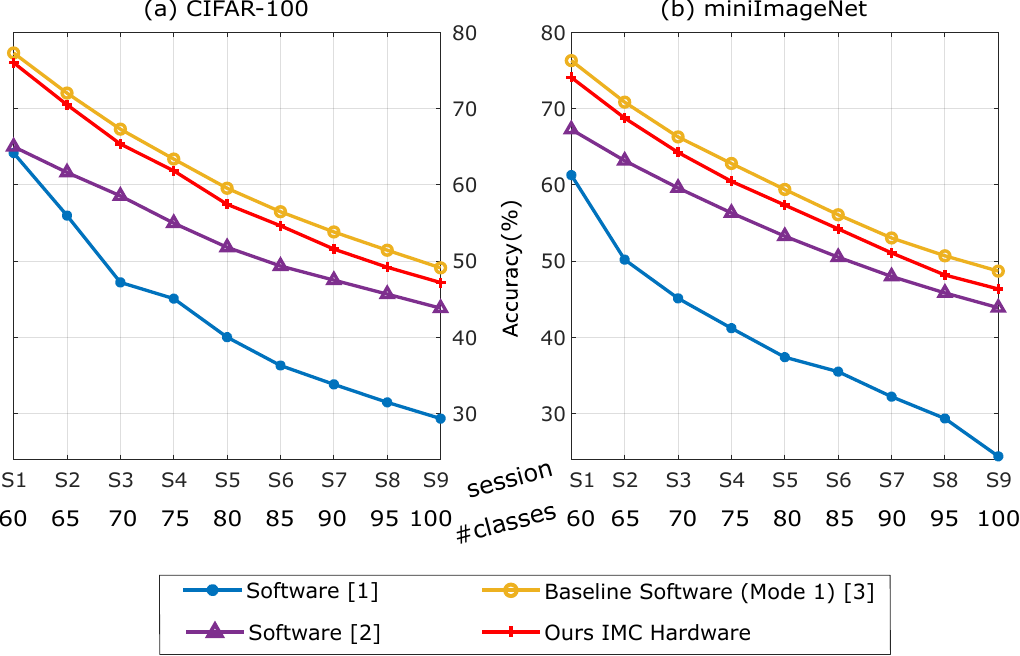}
\caption{FSCL classification accuracy for IMC vs. FP32 software implementations for \textbf{(a)} CIFAR-100 and \textbf{(b)} miniImageNet. As is expected in FSCL, the accuracy progressively reduces with increasing number of sessions, as the learner is progressively exposed to more classes. However, the IMC accuracy is still within 2.5\% of the full-precision software baseline~\cite{C-FSCIL_CVPR22} for all sessions. Notably, our IMC hardware accuracy surpasses other full-precision software methods~\cite{FSCIL_CVPR2020,shi_nips2021}, for all sessions across the two datasets.}
\label{fig:accuracy}
\end{figure}


\section{Experimental Results}
\subsection{Datasets used for evaluation}
\label{sec:dataset}
For the accuracy evaluation, we use the CIFAR-100 and the miniImageNet dataset, restructured to comply with the FSCL setting~\cite{FSCIL_CVPR2020, shi_nips2021,C-FSCIL_CVPR22}. 
Both datasets contain natural images of 100 classes in total, which are divided into a first session (S1) containing 60 classes with 500 training and 100 query examples per class, and eight novel sessions (S2--S9) with 5 novel classes introduced in each session containing 5 support examples and 100 query examples per class.
The updated class vectors occupy 256 rows and 60 columns (classes) on the array in S1, evolving to 100 columns (classes) in the last S9. The conductance evolution of unit cells of the crossbar array corresponding to CIFAR-100 is shown in Fig.~\ref{fig:condmaps2}. 

\subsection{Classification accuracy}
The accuracy obtained for each dataset with our EM on the IMC hardware and various full-precision software baselines are illustrated in Fig.~\ref{fig:accuracy}. Compared to the full-precision software baseline in~\cite{C-FSCIL_CVPR22}, the accuracy degradation with our EM realization is at worst 2.5\% and at best 1.28\% across all sessions. This still makes the accuracy of our EM on the IMC hardware from 2.5\% to 11.01\% higher than the other best performing full-precision software methods reported in~\cite{FSCIL_CVPR2020,shi_nips2021}, considering all sessions across CIFAR-100 and miniImageNet datasets. 

\subsection{Energy estimation}
The energy consumption during incremental class vector updates is estimated using the programming parameters, which include peak pulse current of 150\,uA, flat pulse duration 5\,ns, trailing edge pulse duration 40\,ns and source voltage 2.34\,V. These parameters yield an energy expense of 8.78\,pJ per PCM device during one programming cycle. Considering the vector dimension of 256, the total programming time and energy spent during an incremental update of one class vector are 11.5\,us and 2.25\,nJ, respectively. Given that 25 (5-shots from 5 classes) class vectors are updated during all subsequent sessions (S2--S9), the time and energy spent on updating the class vectors in these sessions are estimated to be 57.6\,us and 56.2\,nJ, respectively.

In comparison, the time and energy spent on similarity search of a single query (including digital to analog conversion, PCM read and analog to digital conversion) are estimated to be 520\,ns and 7.74\,nJ, respectively, during the last session. This leads to a total similarity search time and energy of 5.2\,ms and 77.3\,uJ respectively, as in this session we evaluate 10,000 queries (100 queries per class) in total, compared to just 25 updates of the class vectors. The limited number of updates and the application of SET pulses with low energy ensure that the durability of PCM is not significantly affected~\cite{Y2016tumaNatureNano,Y2020legalloJPD}. 

\renewcommand{\arraystretch}{1.3}
\begin{table}[!ht]
\centering
\caption{Comparison with the related works}\label{tab:results-perception-new}
\resizebox{\linewidth}{!}{
\begin{NiceTabular}{lcccc}
\toprule
& \begin{tabular}[c]{@{}c@{}}Kazemi\\  \textit{~et~al.}~\cite{kazemi_date2021}\end{tabular}  & \begin{tabular}[c]{@{}c@{}}Li \textit{~et~al.}\\  \cite{li_vlsi2021,li_TED2021}\end{tabular} & \begin{tabular}[c]{@{}c@{}}Karunaratne\\  \textit{~et~al.}~\cite{kar_ncom_2021}\end{tabular} & This work \\
 \cmidrule(r){1-1} \cmidrule(r){2-5}
Few-shot learning            &   \checkmark & \checkmark & \checkmark &   \checkmark   \\
Continual learning         &    &  &  &    \checkmark  \\
In-situ accumulation             &  &  &  &    \checkmark  \\
Holographic rep. in EM          & & & \checkmark& \checkmark\\
Analog multibit EM             &  &  &  &    \checkmark  \\
Truly $\mathcal{O}$(1) search\tabularnote{Actual crossbar (no emulation based on individual memory devices) performing all dot product operations in parallel in-memory}            & & \checkmark & &    \checkmark  \\
Query vector dim. ($d$) & 128--192 & 128 & 512 & 256\\
Number of classes            & $\leq$20 & 32 & \textbf{$\leq$100} &    \textbf{$\leq$100}  \\
Similarity search energy\tabularnote{Core energy normalized to one class vector of length 256 ($d=256$)} & - & \textbf{17.5\,pJ} & 25.6\,pJ & 19.1\,pJ\\
Programming energy\tabularnote{per vector element} & - & 99\,pJ & 6240\,pJ & \textbf{8.78\,pJ}\\
Dataset(s)              & Omniglot & Omniglot & Omniglot &    \begin{tabular}[c]{@{}c@{}}miniImageNet\\ \& CIFAR100\end{tabular} \\
\bottomrule
\end{NiceTabular}
}
\label{tab:comparison}
\end{table}
\renewcommand{\arraystretch}{1}

\subsection{Comparison}
We compare features of our work against the related works in Table~\ref{tab:comparison}. Our work shares few-shot learning capability with~\cite{kar_ncom_2021,kazemi_date2021,li_vlsi2021,li_TED2021}, holographic representation of vectors with~\cite{kar_ncom_2021}, and truly $\mathcal{O}(1)$ similarity search capability  with~\cite{li_vlsi2021,li_TED2021}. However, our work is the first to demonstrate continual learning capability using the in-situ accumulation property. This makes our EM unit the fist multibit analog IMC core, whereas in \cite{li_vlsi2021,li_TED2021} storage and queries use binary vectors. Furthermore, our EM unit can handle up to 100 class vectors for the natural image datasets, making it the largest truly $\mathcal{O}(1)$ similarity search engine with analog states to date.

In Table~\ref{tab:comparison}, we also compare energy saving we gain by programming the support vectors using the in-situ accumulation instead of programming them from scratch in the novel bit lines. Programming with the in-situ accumulation is at least 4.7$\times$ energy efficient than the normal programming, because the in-situ accumulation requires a SET pulse of shorter duration. Our similarity search energy remains on par with other works.

\section{Conclusion}
We present a hardware based on an in-memory compute core consisting of PCM devices to implement the EM operations required in FSCL. We demonstrate for the first time how support vectors are accumulated in-situ using the progressive crystallization property of PCM. The proposed approach leads to physically superposed representations and enhanced energy savings, while retaining the accuracy within 2.5\% from the full-precision software baseline. 

\section*{Acknowledgements}
This work is supported by the IBM Research AI Hardware Center, and the Center for Computational Innovation at Rensselaer Polytechnic Institute for computational resources on the AiMOS Supercomputer.

\bibliographystyle{IEEEtran}
\tiny
\bibliography{refs}

\end{document}